# Two Stage Segmentation of Cervical Tumors Using PocketNet

Awj Twam[1], Adrian E. Celaya[1], Megan C. Jacobsen[1], Rachel Glenn[1], Peng Wei[4], Jia Sun[4], Ann Klopp[2], Aradhana M. Venkatesan[3], David Fuentes[1]

[1]Department of Imaging Physics, The University of Texas MD Anderson Cancer Center, Houston, Texas, USA

[2]Department of Radiation Oncology, The University of Texas MD Anderson Cancer Center, Houston, Texas, USA

[3]Department of Abdominal Imaging, The University of Texas MD Anderson Cancer Center, Houston, Texas, USA

[4]Department of Biostatistics, The University of Texas MD Anderson Cancer Center, Houston, Texas, USA

**Abstract.** Cervical cancer remains the fourth most common malignancy amongst women worldwide.[1] Concurrent chemoradiotherapy (CRT) serves as the mainstay definitive treatment regimen for locally advanced cervical cancers and includes external beam radiation followed by brachytherapy.[2] Integral to radiotherapy treatment planning is the routine contouring of both the target tumor at the level of the cervix, associated gynecologic anatomy and the adjacent organs at risk (OARs). However, manual contouring of these structures is both time and labor intensive and associated with known interobserver variability that can impact treatment outcomes. While multiple tools have been developed to automatically segment OARs and the high-risk clinical tumor volume (HR-CTV) using computed tomography (CT) images,[3,4,5,6] the development of deep learning-based tumor segmentation tools using routine T2-weighted (T2w) magnetic resonance imaging (MRI) addresses an unmet clinical need to improve the routine contouring of both anatomical structures and cervical cancers, thereby increasing quality and consistency of radiotherapy planning. This work applied a novel deep-learning model (PocketNet) to segment the cervix, vagina, uterus, and tumor(s) on T2w MRI. The performance of the PocketNet architecture was evaluated, when trained on data via five-fold cross validation. PocketNet achieved a mean Dice-Sorensen similarity coefficient (DSC) exceeding 70% for tumor segmentation and 80% for organ segmentation. Validation on a publicly available dataset from The Cancer Imaging Archive (TCIA) demonstrated the model's robustness, achieving DSC scores of 67.3% for tumor segmentation and 80.8% for organ segmentation. These results suggest that PocketNet is robust to variations in contrast protocols, providing reliable segmentation of the regions of interest.

## Introduction

In 2022, the Global Cancer Observatory documented 662,301 cases of uterine cervical cancer worldwide, ranking it as the 9th leading cause of cancer-related mortality, with 348,874 recorded deaths.[7] Cervical cancer primarily occurs due to the integration of human papillomaviruses (HPV) in the host genome.[8] While cervical cancer remains a largely preventable disease through HPV vaccination and routine screening, the incidence and mortality rates of cervical cancer have not changed in the past twenty years.[9] Diagnosis and treatment of cervical cancers heavily rely on imaging, with magnetic resonance imaging (MRI) as the primary modality for staging and evaluation of locally advanced cervical cancers (LACCs) due to its superior soft tissue contrast as compared to other clinical imaging modalities, such as ultrasound or CT.[10] The main form of treatment for locally advanced cervical cancer is definitive radiation therapy (RT), accompanied by cisplatin-based chemotherapy. Definitive RT for cervical cancer includes external beam RT and brachytherapy (BT), both of which necessitate manual segmentation of the cervical tumor before therapy. Although an essential part of treatment, manual segmentation can be a time-consuming task and contribute to a delay in the start of BT following applicator insertion. Automated segmentation, if available, could facilitate adaptive planning, which requires daily re-contouring to customize the treatment plan to daily changes in anatomy due to movement of bowel and bladder or to account for the tumor volume reduction that occurs over the course of treatment. Although adaptive planning has been demonstrated to reduce radiation doses to normal tissues, it is not widely employed due to the time intensive process of manual segmentation. Manual segmentation is also subject to significant inter- and intraobserver variability[11], which limits researchers' ability to train deep-learning models using large datasets. Automatic segmentation addresses these challenges and presents an opportunity for advancement in the development of multi-institutional models for cervical cancer outcomes. Beyond segmentation, DL models, particularly convolutional neural networks (CNNs), have shown great potential in the diagnosis of uterine cervical cancer. CNNs applied to T2-weighted MR images of the female pelvis have demonstrated high accuracy (90.8%) when determining the binary presence or absence of cervical cancer, which was comparable to the performance of experienced readers.[12] Building on this success, CNNs have emerged as a powerful class of DL models, further advancing the capabilities of medical image analysis.

One of the most popular fully CNN architectures for medical image segmentation is the U-Net. The U-Net's success in medical image segmentation is largely due to a combination of feature extraction and reconstruction, which enables it to accurately delineate complex anatomical structures. However, the U-Net's architecture is also resource-intensive, requiring a significant number of parameters, which can make it challenging to deploy in environments with limited computational resources. Therefore, we propose using a PocketNet approach to segment the uterus, vagina, cervix and tumors. PocketNet builds off the U-Net backbone but dramatically reduces the number of parameters to minimize computational burden while producing similar performance to that of full-sized networks.[13] This work evaluated the performance of a two-stage PocketNet approach for image segmentation of cervical cancers on T2w MRIs in patients treated with definitive RT. To validate the robustness of the model, we utilized a publicly available dataset from The Cancer Imaging Archive (TCIA).

## Materials and Methods

Within our institution, we collected over 300 pre-treatment T2w MR images from 102 patients diagnosed with cervical cancer who ultimately underwent definitive chemoradiotherapy. Two volumetric contours were manually segmented for each patient: 1) the combined uterus, cervix, vagina, and tumor, collectively labeled as 'organ' and 2) the cervical tumor alone. Manual segmentations were performed by trained staff and approved by a fellowship-trained abdominal radiologist with 22 years' experience and were used to define ground truth. The initial PocketNet architecture was trained on 35 patients for segmentation of the 'organ'. A second PocketNet model was trained on 102 patients including the organ mask and tumor, with the requirement that the final tumor segmentation was contained within the organ (Figure 1). Dice similarity coefficients (DSC) and 95th Percentile Hausdorff distance (Haus95) were calculated for each contour as a measurement of the accuracy and maximum distance between ground truth and automated contours, respectively. In addition to the institutional dataset, TCIA's publicly available Cervical Cancer Tumor Heterogeneity (CCTH) collection provides a useful resource by offering multi-parametric imaging data (T1- and T2-weighted, DCE, DWI MRI, and FDG PET/CT) obtained from patients with advanced cervical cancer at three treatment points. This dataset, collected using standardized protocols, included 18 gold-standard tumor contours based on T2-weighted MRI.

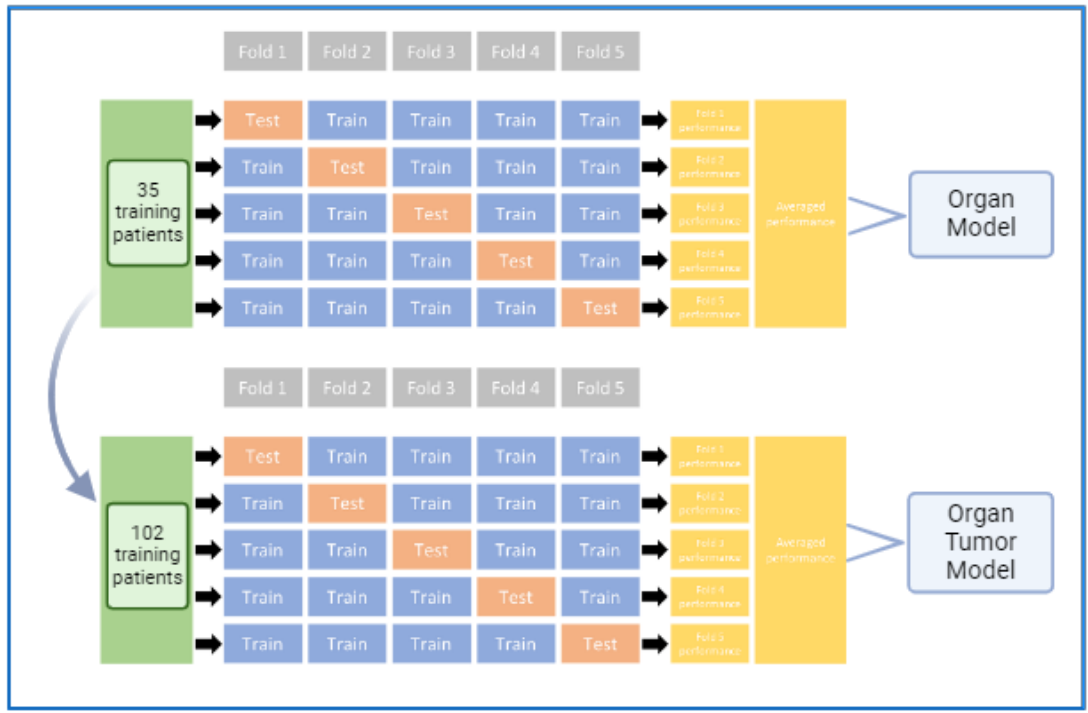

**Figure 1.** Overview of five-fold cross validation training process

**Data:** Preparation of imaging data is essential to ensure accurate and reproducible predictions. Inclusion criteria for MR images used for training, validation, and testing were as follows: 1) Visibility of the entire cervix, vagina, uterus, and tumor within the image; 2) presence of entire organ segments; 3) acceptable image quality such that the boundary organ and tumor was identifiable without the aid of a pre-existing contour. Images were manually inspected, and quality assurance was conducted over the entire dataset to confirm MR images met these criteria. The pre-processing pipeline from the Medical Imaging Segmentation Toolkit (MIST) was used to prepare the data. This pipeline processes NIfTI files and outputs reoriented, resampled, windowed, and normalized NumPy arrays. To maintain consistency with the model's training data, images acquired from the publicly available TCIA dataset, initially in DICOM format, underwent conversion to NIfTI. Each image was then resampled and reoriented from the sagittal view to an axial view in the right-anterior-inferior orientation. This pre-processing step was necessary to standardize orientation and spacing across all datasets. Table 1 details the patch size and pixel spacing of data included in the analysis. All MR images were prepared using a patch size of 256 x 256 x 32 and a pixel spacing of 0.469 x 0.469 x 5 mm. The patch size was derived from the median resampled image size by selecting the nearest power of two less than or equal to each dimension, constrained by a maximum patch size.

**Pre-processing:**

| Data | Patch Size | Pixel spacing (mm) |
|------|-----------|--------------------|
| MD Anderson | 256 x 256 x 32 | 0.469 x 0.469 x 5 |
| TCIA | 256 x 256 x 32 | 0.469 x 0.469 x 5 |

**Table** 1. Pre-processing of data

**Network architecture:** In a PocketNet architecture, the number of feature maps resulting from each convolution layer remains constant regardless of spatial resolution (Figure 2). Traditional CNNs double the number of features when going from higher to lower resolutions. With the use of PocketNet, this results in faster training and inference times while simultaneously lowering memory usage and requirements.

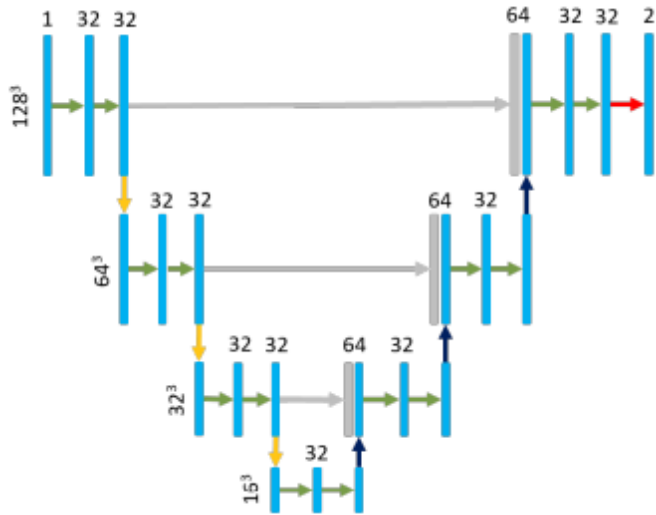

**Figure 2.** PocketNet architecture. Briefly, our network is constructed from a composition of convolution and down sampling operations that extract features along a contracting path. Similarly, an expanding path consists of convolution and up-sampling operations with 'long skip' connections to integrate features from the corresponding down sampling operations. Four resolution levels are used. At a given resolution, the feature-maps of all preceding layers are used as inputs, and its own feature-maps are used as inputs into all subsequent layers. The key difference in our architecture is that the number of layers is not heuristically increased at each resolution. Each convolutional operation uses a 3x3 kernel size and is followed by a batch normalization and a ReLU activation function.

**Training protocols and hyperparameters:** We used an Nvidia Quadro RTX 8000 graphical processing unit (GPU) with a batch size equal to the number of patients included in each fold, trained with 1,000 epochs. During training, network weights were determined using the ADAM optimizer with a learning rate set to 0.0001 and use of a cosine scheduler. Additionally, automatic mixed precision was used during training to reduce the time and memory requirements. Classification tasks include the Dice with categorical cross-entropy as our loss function. Regularization features include deep supervision and segmentation tasks using the $L_2$ norm. In addition, residual blocks were also utilized in the architecture to improve training by resolving the degradation problem with the use of skip-connections around convolution layers. Finally, we postprocess predictions by taking the largest connected component, which eliminates noise and small islands distant to the body of the contour. To evaluate the validity of a predicted segmentation mask, the DSC and Haus95 are used as outcome metrics to evaluate segmentation accuracy. In addition to the parameters described above, performance of PocketNet with default parameter settings was conducted separately to compare results. The results revealed that inclusion of auxiliary parameters described above, highlighted in Appendix A, and included in our training model increased average dice scores by 5.04% for organ and 7.95% for tumor as opposed to the default settings.

## Experimental Design

Using the data and model described above, we performed a two-stage segmentation, outlined in Figure 3, using PocketNet to evaluate the architecture's performance for T2w MRI-based automated segmentation of the cervical tumor and gynecologic organs.

First, we performed a five-fold cross-validation on the initial training dataset of 35 cancer patients to generate an organ model containing the tumor. Inference was then performed to generate organ prediction masks on 102 patients. Then, we combine the organ and tumor masks manually segmented into the same dataset of 102 patients. We perform another five-fold cross-validation on this dataset which includes the T2-weighted image and mask (organ and tumor combined).

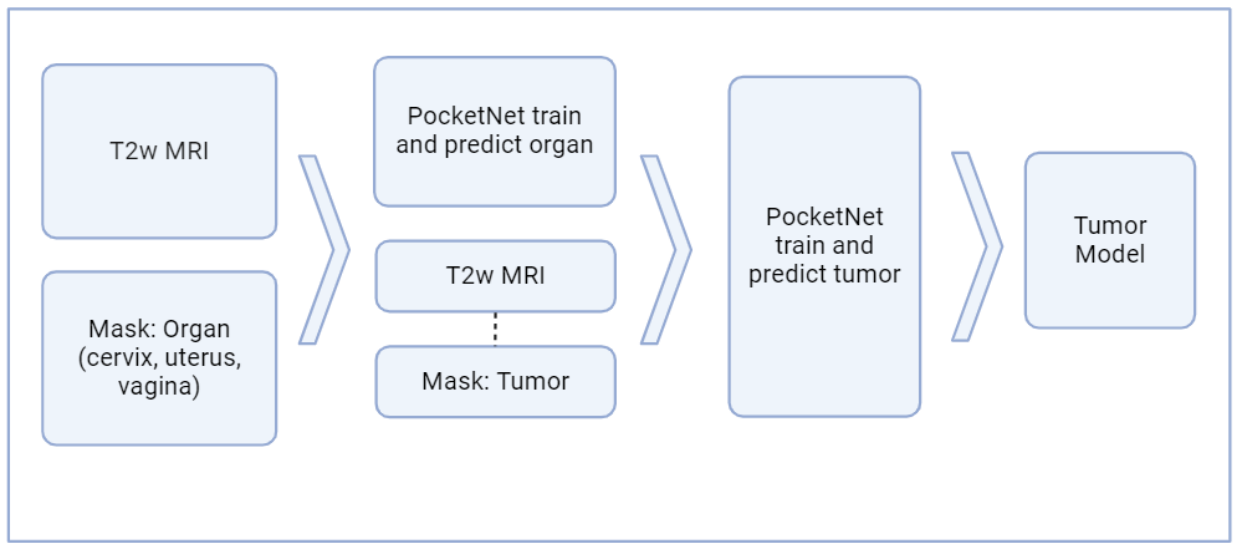

**Figure 3.** Overview of two-stage segmentation process. The first stage segments the organ. The second stage requires that the tumor segmentation be contained within the organ segmentation from the first stage.

Finally, to evaluate the generalizability of our PocketNet model, we applied it to an externally available dataset provided by TCIA. The PocketNet model trained on the institutional dataset (102 cases) was applied to the images provided by TCIA. Inference was performed on these 18 T2w MRIs, which generated automated segmentations for both the gynecologic organs and cervical tumors.

**Ethics approval:**

All data used in this study was collected and analyzed under an approved Institutional Review Board protocol (IRB).

## Results

**Quantitative evaluation:**

**Table 2**. Validation metric means and standard deviations for PocketNet.

|  | Dice-Sorenson coefficient | | Hausdorff distance 95th percentile | |
|---|---|---|---|---|
| **Dataset** | **Mean** | **Standard deviation** | **Mean** | **Standard deviation** |
| **MDACC** | | | | |
| Organ | 80.98% | 12.24% | 64.23 mm | 65.85 mm |
| Tumor | 71.27% | 20.87% | 38.59 mm | 56.28 mm |
| **TCIA** | | | | |
| Organ | 80.78% | 10.6% | 14.76 mm | 7.45 mm |
| Tumor | 67.35% | 22.76% | 15.95 mm | 13.08 mm |

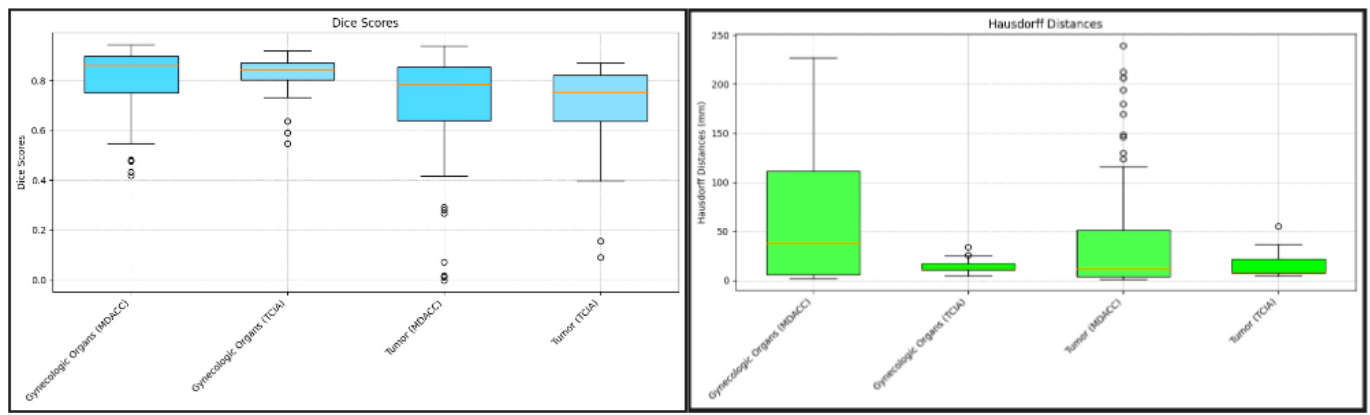

**Figure 4.** Boxplots of Dice Sorensen Coefficient and Hausdorff Distances for Gynecologic Organs and Tumor

Table 2 and Figure 4 summarize the performance of the segmentation model across two datasets (MDACC and TCIA) using DSC and Haus95 as evaluation metrics. An average DSC of 81% was observed for organ segmentation on the MDACC dataset (with a standard deviation of 0.122) and an average DSC of 71% was observed for tumor segmentation (with a standard deviation of 0.209). Additionally, the average Haus95 was measured at 64.23 mm for organs (with a standard deviation of 65.85 mm) and 38.59 mm for tumors (with a standard deviation of 56.28 mm). These results provide a quantitative overview of the model's performance while the standard deviations highlight the variability within each metric, indicating certain cases deviated more from the mean, particularly in boundary accuracy. Figure 4 visualizes these metrics using box and whisker plots, offering a more detailed representation of the variability and distribution of the segmentation accuracy and boundary discrepancies across cases. The median DSC for organ segmentation was relatively high at 0.86, indicating that the majority of cases have a good agreement between the predicted and ground truth segmentations. The interquartile range (IQR) for organ DSC is comparatively narrow, suggesting that most of the DSC values are clustered around the median. The upper quartile reaching 0.90 reflects that many cases perform exceptionally well. However, the lower quartile at 0.75 suggests that some cases have moderate performance, with four outliers below the lower quartile, highlighting cases where segmentation performance was particularly poor. In comparison to organ DSC, median DSC for tumor is around 0.79, reflecting that tumor segmentation is generally more challenging. The boxplot for tumor DSC shows a wider IQR, indicating more variability in segmentation performance across different cases. The 25th percentile at 0.64 further emphasizes this variability. Again, six outliers below the lower quartile emphasize a need for improvement in segmentation accuracy and performance. The median Haus95 value for organ segmentation is 37.87 mm, but the mean is higher at 64.23 mm. Distribution of Haus95 for organ is broad, reflecting a substantial variability in the boundary accuracy, without presence of outliers. The median Haus95 value for tumor segmentation is lower at 11.7 mm, indicating a better boundary accuracy compared to organ. Although the IQR is narrower, depicting more consistent performance in tumor boundary predictions, the presence of a relatively high mean (38.59 mm) suggests a few cases with large errors, validated with the presence of outliers above the upper whisker.

The box plots for DSC reveal that on the institutional dataset, organ segmentation generally performs better and more consistently than tumor segmentation, with fewer outliers and a narrower distribution of values. Conversely, the Haus95 box plots highlight the variability and challenges in boundary accuracy, particularly for the organ segmentation, where a few cases deviate from the norm. Tumor segmentation shows a more consistent performance but still exhibits some challenging cases with substantial boundary errors.

Analysis of the TCIA dataset reveals several important insights about the segmentation model's performance in comparison to the MDACC dataset. In contrast to the MDACC dataset, tumor segmentation performed on the TCIA dataset demonstrates a more consistent DSC range, with a higher median (0.75) and fewer outliers than organ segmentation, suggesting that the model generalized well for organ segmentation. Segmentation performed on the TCIA dataset achieved a mean DSC of 67.35%, slightly lower than that of the MDACC dataset (71.27%). However, there remain two outliers in the tumor DSC scores, corresponding to cases where segmentation accuracy is lower. These challenging cases involved smaller and more poorly defined tumors for which the boundaries were less distinct, making accurate segmentation more difficult. Organ segmentation performed on the TCIA dataset achieved a high mean DSC of 80.78%, closely matching segmentation performed on the MDACC data set (80.98%). However, segmentations performed on the TCIA dataset demonstrated a narrower standard deviation (10.6% vs 12.24%), indicating more consistent performance despite its smaller size. The median Dice score for organ segmentation using the TCIA dataset was 0.85, with a relatively narrow IQR, reflecting minimal variability and reliable model performance across its limited cases. In terms of boundary accuracy of organ segmentation, TCIA achieves a lower mean Haus95 (14.76 mm) compared to MDACC (64.23 mm), with a narrower range of variability. This highlights that TCIA organ segmentations were not only accurate, but also demonstrated better boundary consistency than segmentations performed on the MDACC dataset, which had a higher variability likely due to the larger and more diverse dataset.

In summary, the TCIA dataset demonstrates the model's generalizability and robustness as a validation set for the MDACC-trained model. The robust performance of segmentations using the TCIA dataset, particularly in terms of boundary accuracy and consistent Dice scores, highlights the model's ability to adapt to new datasets. While organ segmentation generalizes exceptionally well, tumor segmentation remains more challenging but still shows reliable performance.

**Qualitative evaluation:**

Figure 5 presents two examples of automatic contours that performed well. Representative segmentations of organ and tumor (5a-c) demonstrated Dice score of 0.917 for organ and 0.883 for tumor, indicating a high degree of overlap between the automatically generated contours (C) and the ground truth (B). 5d-f demonstrates organ and tumor segmentations in a different patient, illustrating even better performance for the organ, with a Dice score of 0.946, and a comparable score of 0.889 for the tumor. These high Dice scores reflect the algorithm's strong capability to accurately delineate both the organ and tumor in these cases. In contrast, Figure 6 displays instances where the auto-segmentation algorithms did not perform as well. Figure 6a-c had a dice score of 0.786 for organ and 0.528 for tumor, indicating a noticeable decrease in accuracy, particularly for the tumor. Figure 6d-f had a Dice score of 0.550 for the organ and a low score of 0.018 for organ and tumor. The contrast between the results in Figures 5 and 6 emphasizes the importance of continuous refinement and validation of automatic segmentation algorithms.

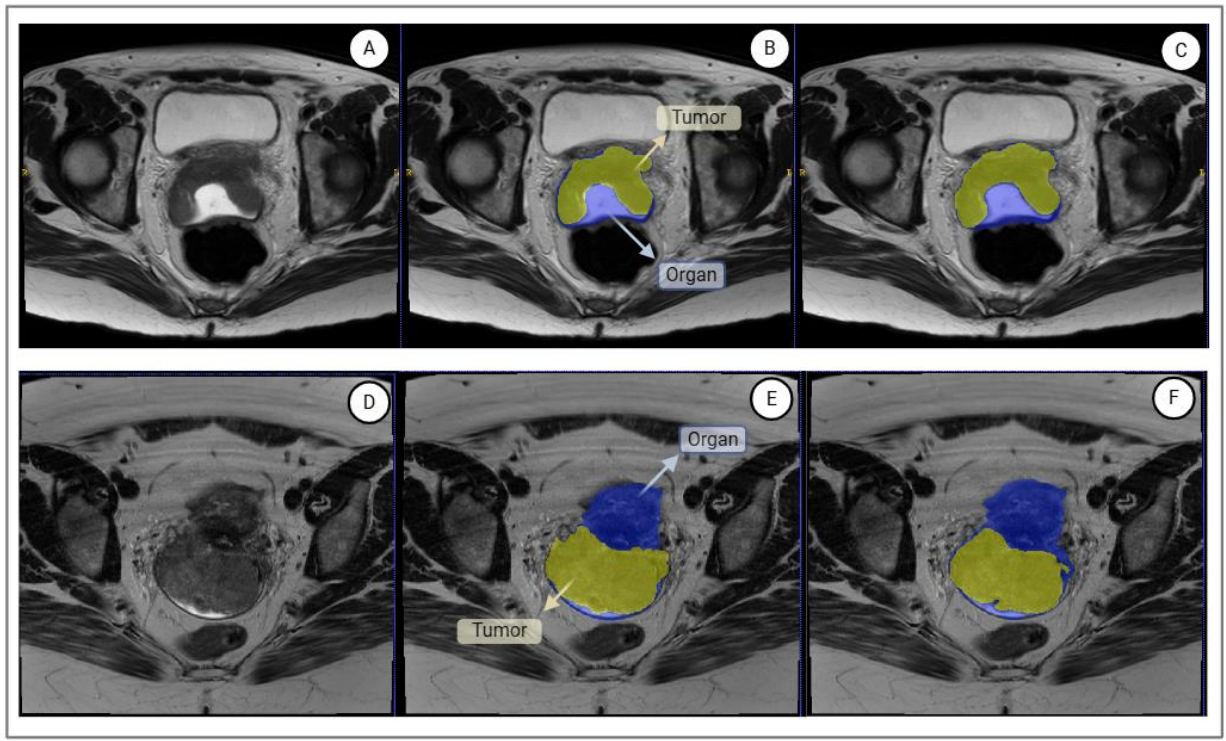

**Figure 5.** Manual and automated segmentation in patients. A) Axial T2-weighted image of patient with FIGO stage IVA squamous cell carcinoma of the cervix. D) Axial T2-weighted image of patient with FIGO stage IIB adenocarcinoma of the cervix. B, E) Manual segmentation of organ (blue) and tumor (yellow). Note that the tumor is contained within the gynecologic organ contour is inclusive of the tumor to enable the two-step segmentation model. C, F) Automated segmentation of organ (blue) and tumor (yellow). The manual and automated tumor contours demonstrate strong agreement with DSC of 0.883 and 0.889, respectively.

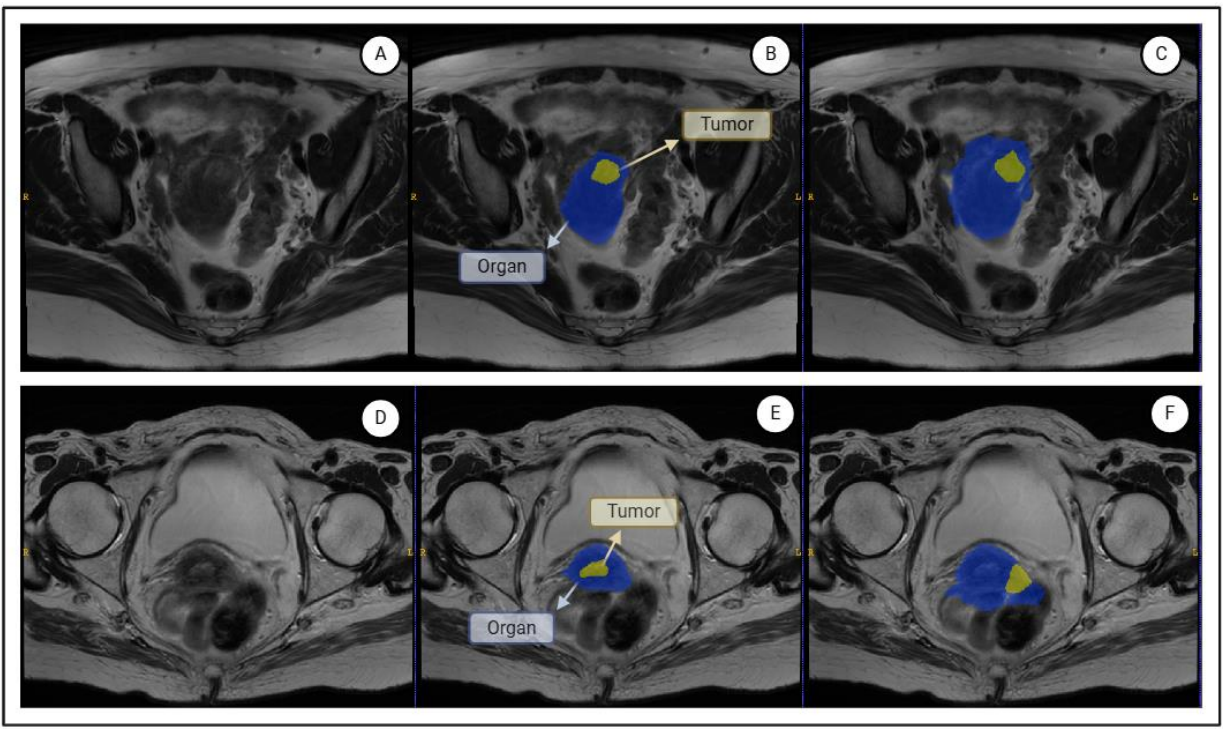

**Figure 6.** Results of segmentation in two patients with FIGO Stage IIIC1cervical cancer. A, D) Axial T2-weighted image B, E) Manual segmentation of organ (blue) and tumor (yellow). C, F) Autosegmentation of organ (blue) and tumor (yellow). The segmentations demonstrate poor agreement in the inferior aspect of the tumor and organ, confounded by the proximity between the cervix and the adjacent rectosigmoid colon.

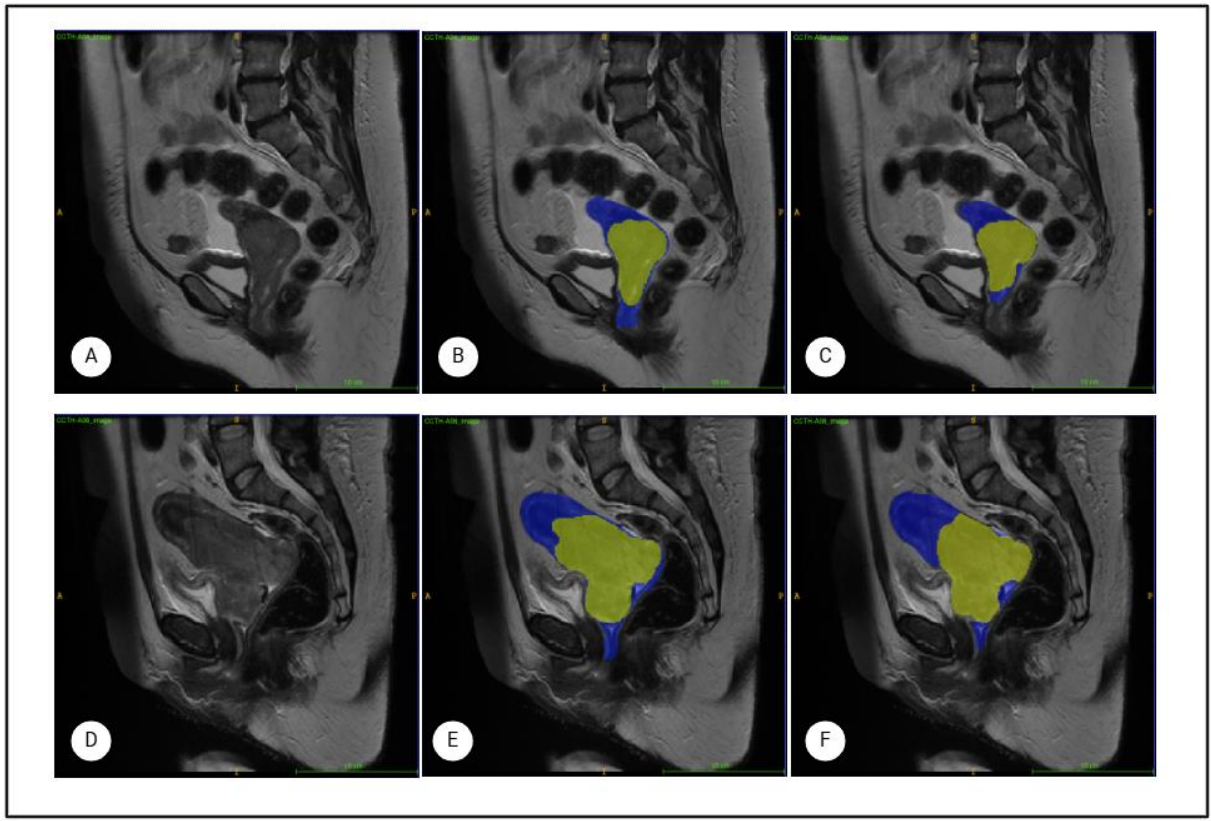

**Figure 7.** Segmentation model applied to the TCIA dataset. Panels A and D display sagittal T2 Weighted MRI scans containing the organ and tumor. Panels B and E illustrate the manual segmentations with the organ highlighted in blue and tumor in yellow. Panels C and F present the autosegmentations generated by PocketNet, using the same color scheme. The manual and automated segmentations show good agreement with DSC values of 0.87 (C) and 0.92 (F) for organ segmentation and 0.857 (C) and 0.87 (F) for tumor segmentation.

Figure 7 provides a visual representation of the segmentation model's performance when applied to the TCIA dataset, demonstrating its ability to accurately identify and delineate both gynecologic organs and cervical tumors. The manual and automated segmentations achieved DSC values of 0.87 and 0.92 for segmentation of gynecologic organs and 0.857 and 0.87 for tumor segmentation. These high values highlight the model's strong performance in capturing tumor boundaries. Both examples underscore PocketNet's ability to reproduce accurate segmentation results.

## Discussion

Accurate and reliable identification of these anatomical regions and tumors is critical for improving accuracy during diagnosis and radiotherapy treatment planning. A study conducted by Chung et al. demonstrated the practical impact of DL based autosegmentation systems in clinical workflows.[14] The results showed that using an autosegmentation system reduced the average contouring time by approximately half an hour. Among the participating physicians, 84% reported satisfaction with the autosegmentation model, considering it to be helpful in clinical practice.[14] Moreover, the widespread acceptance of autosegmentation tools across multiple institutions suggests the potential for these models to become standard features in radiotherapy planning. The primary objective of this study was to develop and apply a deep learning model for the robust segmentation of cervical, uterine, vaginal, and tumor tissues with T2w MRI. To achieve this, we employed the PocketNet model, a compact yet powerful deep learning architecture designed for efficient image segmentation tasks. The PocketNet paradigm introduces several optimizations that make it particularly well-suited for resource-constrained environments while maintaining high performance levels. One of the most notable advantages of PocketNet is its reduced GPU memory usage while decreasing inference times compared to a traditional U-net. Another significant benefit of PocketNet is the reduction in training time. Celaya et al. evaluated the performance of PocketNet in a variety of medical imaging datasets, comparing its efficiency and effectiveness to that of full-sized architectures like U-Net. These results demonstrate that the PocketNet architecture reduces memory usage and speeds up training time for every batch size.[13]

The performance of automatic segmentation algorithms in cervical cancer treatment presents both opportunities and challenges, as evidenced by the varying results highlighted in Figures 5 and 6. A key observation from these results in the apparent relationship between tumor size and segmentation accuracy. Larger tumors, as seen in Figure 4, tend to be segmented more accurately, resulting in higher Dice scores, while smaller tumors, like those in Figure 5, often yielded lower Dice scores. Smaller tumors may be entirely confined to the cervix or lower uterine segment with less signal difference between the normal tissue and tumor on T2w imaging. In contrast, larger masses are more likely to involve the full thickness of the cervix or have exophytic components that are relatively simple to differentiate from the surrounding normal tissue. This finding underscores the need for further refinement and validation of these models, such as the inclusion of diffusion-weighted MRI to identify regions of restricted diffusion that are specific to cervical tumors.

Recent advancements in deep learning have brought about significant improvements in the automatic segmentation of medical images, including that of cervical cancer. In a recent study conducted by Kano et al., a deep learning model applied to DWI achieved a mean DSC of 0.77 for cervical tumor segmentation.[15] This result demonstrates the robustness of the PocketNet model as compared to existing DL methods. The comparable performance of our T2w-based model, which achieved a mean DSC of 0.71, to this DWI-based model suggests that PocketNet has the potential to achieve high accuracy across various imaging sequences with further development. Another study comparing the segmentation performance of a DL algorithm (specifically, a 3D U-Net) with manual segmentations provided by two radiologists (R1 and R2) reported median DSCs of 0.60 for R1 and 0.58 for R2.[16] In addition, a study utilizing a U-Net based triple-channel model using sagittal diffusion-weighted MRI (DWI) with b-values of 0 and 1000 s/mm$^2$ and apparent diffusion coefficient (ADC) maps produced a promising average DSC of 0.82, further demonstrating the capability of DL models to enable accurate segmentation of cervical cancer.[17] One of the most promising applications of DL models in cervical cancer treatment is their use in developing daily online dose optimization strategies. This application of DL models has the potential to eliminate the need for internal target volume expansions, ultimately resulting in the reduction of toxicity events related to intensity modulated radiation therapy.[18]

Despite the promising results obtained from this institutional dataset, this study has limitations, including the size of this dataset and the lack of multi-center training data. To further improve segmentation accuracy, expanding the training dataset to include a larger and more diverse set of images is an essential next step. By incorporating a more diverse dataset, the model can be trained on a broader range of tumor sizes, morphologies, and degrees of soft tissue contrast, leading to more consistent and accurate segmentations. Integrating additional imaging sequences such as DWI may also enhance the model's robustness. In addition, ensuring that these diverse datasets are collected using consistent, standardized imaging protocols will minimize variability and improve the model.

The validation process using TCIA's CCTH data provided critical insights into the generalizability of PocketNet. Despite differences in imaging protocols, institutions, and patient populations, the model maintained high segmentation accuracy, achieving mean DSC values of 80.8% for gynecologic organ segmentation and 67.3% for cervical tumor segmentation. This external validation has confirmed that the model can accurately segment cervical, uterine, and vaginal tissues, as well as tumors, on T2-weighted MRI.

In conclusion, this preliminary work employing PocketNet to segment the cervix, uterus, vagina, and tumor(s) on T2w MRI in patients undergoing definitive RT for cervical cancer demonstrated reliable segmentation of both tumor and the gynecologic organs, suggesting the utility of this time efficient approach to automated segmentation. Further work will refine the performance and validate this approach, with the goal of employing this methodology to support routine clinical practice. Continued advancements in data harmonization, model training, and cross-validation will be key drivers to achieve this goal, ultimately leading to greater standardization of definitive radiotherapy and improved clinical outcomes for patients with cervical cancer.

## Conclusion

Although essential for cervical cancer treatment planning, manual segmentation of the cervix, uterus, vagina, and tumors is a relatively time-consuming process that is subject to variability. Our proposed deep learning segmentation model, PocketNet, with its compact architecture and optimizations for reduced GPU memory usage and faster inference times, addresses these challenges, by providing an approach that automates the segmentation of anatomical regions of interest for cervical cancer radiation treatment planning. These results demonstrate that PocketNet achieved an average DSC of 81% for organ segmentation and 71% for tumor segmentation. The CCTH collection not only strengthens our model's validation but also serves as a rich resource for ongoing research. Future work employing this model is anticipated to further refine and validate this approach for automated segmentation.

## Data Availability

The MR images from MD Anderson are not publicly available at this time.

## Acknowledgements

This work was supported through the QIAC Partnership in Research (QPR) Program, Multidivisional Shark Tank, and the MD Anderson Strategic Research Initiative Development (STRIDE) Program – Tumor Measurement Initiative. Support under NIH R01CA195524 and NSF Award NSF-2111147 is gratefully acknowledged.

## Competing Interests

The authors declare no competing interests.

## Appendix

### Appendix A: Training Protocols and Hyperparameters

**Table A1.** Comparison of default and customized PocketNet settings used during training

| Setting | Default | Customized |
|---|---|---|
| **Optimizer** | ADAM | ADAM |
| **Learning Rate** | 0.0003 | 0.0001, lr-scheduler cosine |
| **Precision** | AMP | AMP |
| **Loss Function** | Dice with Cross Entropy Loss | Dice with Cross Entropy Loss |
| **Regularization Features** | None | Deep-supervision, L2-reg |